\documentclass{article}

\usepackage{arxiv}

\usepackage[utf8]{inputenc} 
\usepackage[T1]{fontenc}    
\usepackage{hyperref}       
\usepackage{url}            
\usepackage{booktabs}       
\usepackage{amsfonts}       
\usepackage{nicefrac}       
\usepackage{microtype}      
\usepackage{lipsum}
\usepackage{graphicx}
\graphicspath{ {./images/} }
\usepackage{amsmath,amsfonts}
\usepackage{algorithmic}
\usepackage{multirow}
\usepackage{algorithm}
\usepackage{array}
\usepackage[caption=false,font=normalsize,labelfont=sf,textfont=sf]{subfig}
\usepackage{textcomp}
\usepackage{stfloats}
\usepackage{url}
\usepackage{verbatim}
\usepackage{graphicx}
\usepackage{cite}
\usepackage{comment}
\usepackage{tablefootnote}
\usepackage{placeins}

\usepackage{svg} 

\title{Beyond Single-Axis Testing: Paired Evaluation of Compound Robustness in Vision-Language-Action Policies}

\author{
 Hiroki Sawada \\
  Sony Computer Science Laboratory\\
  Tokyo, Japan \\
  \texttt{hiroki.sawada@csl.sony.co.jp} \\
   \And
 Shunichi Kasahara \\
  Sony Computer Science Laboratory\\
  Tokyo, Japan \\
  \texttt{kasahara@csl.sony.co.jp} \\
}

\begin{document}
\maketitle
\begin{abstract}
Vision-language-action policies are typically evaluated one perturbation at a time, providing a useful diagnosis of their sensitivity to individual distribution shifts.
Real-world deployment, however, may involve several shifts simultaneously, and it remains unclear how these individual robustness measurements compose.
We ask whether compound robustness can be inferred from single-axis evaluations. 
We introduce LIBERO-CTRL, a six-axis benchmark that pairs each initial state across single-axis conditions and a matched simultaneous condition.
This design reveals two opposing outcome changes that aggregate success rates cannot distinguish: emergent failures, where all single-axis rollouts succeed but the simultaneous rollout fails, and compensated successes, where at least one single-axis rollout fails but the simultaneous rollout succeeds. 
Because one transition decreases compound success while the other increases it, they can cancel, making aggregate compound performance appear consistent with single-axis measurements even when individual outcomes differ substantially. 
These opposing transitions can largely cancel in aggregate: even when the difference between the two transition rates is not statistically distinguishable from zero, as many as 29.0\% of matched initial states still change outcome. 
Across six policies and three severity levels, such outcome changes reach 34.5\% in the most affected condition.
The relative prevalence of the two transitions varies across policies and severities, while the transition rates remain similar under independent re-evaluation of stochastic policies.
Compound robustness therefore cannot be characterized from aggregate single-axis success rates alone; matched per-instance evaluation is needed to reveal how joint perturbations alter behavior.
\end{abstract}


\section{Introduction}
\label{sec:intro}

Vision-language-action (VLA) policies are commonly evaluated by aggregate success rates under standard benchmark conditions~\cite{liu2023libero,mees2022calvin}, while robustness evaluations often characterize individual sources of distribution shift separately~\cite{liberoplus2025,liberopro2025}.
Such single-axis evaluations are useful because they reveal which aspects of the observation-action loop a policy is sensitive to.
Real-world deployment, however, exposes a policy to several shifts simultaneously, and it remains unclear how robustness to individual perturbations relates to robustness when multiple perturbations occur together.
Directly evaluating the compound condition reveals its overall success rate, but not whether compound failures are already explained by vulnerabilities visible under the constituent perturbations or arise only when those perturbations are combined.
We therefore ask whether, and in what sense, compound robustness can be inferred from single-axis evaluations.

Aggregate single-axis success rates alone cannot answer this question because they discard which initial states succeed or fail.
For example, two perturbations may each yield $75\%$ success rate among different initial states while failing on exactly the same $25\%$ of them, or on largely different ones.
The marginal success rates are identical, but the two cases imply very different overlap in which initial states are robust to both perturbations individually.
Paired evaluation of the same initial states reveals this cross-axis structure, but still does not determine what happens when the perturbations are applied together.
An initial state may succeed under every perturbation individually but fail when those same perturbations are combined; conversely, one that fails under one or more individual perturbations may succeed under the combined condition.
We refer to these two transitions as \emph{emergent failures} and \emph{compensated successes}, respectively.
Because they affect aggregate success in opposite directions, they can cancel, making compound performance appear consistent with the single-axis results even when many matched initial states change outcome.

We introduce LIBERO-CTRL, a fully paired six-axis perturbation protocol that links individual and simultaneous perturbation outcomes at the level of matched initial states.
For each initial state, LIBERO-CTRL evaluates each perturbation individually and then applies the same perturbation values simultaneously.
This design separates two questions that aggregate evaluation conflates: whether vulnerabilities to different single-axis perturbations occur on the same initial states, and whether applying those perturbations together changes outcomes beyond what is already observed individually.

We apply this protocol to six language-conditioned VLA policies spanning sub-million to multi-billion parameter scales and three perturbation severity levels.
The resulting paired evaluations reveal substantial bidirectional outcome changes.
Even when emergent failures and compensated successes nearly cancel in aggregate, as many as $29.0\%$ of matched initial states still change outcome.
Across all policy-severity conditions, these changes reach $34.5\%$ in the most affected condition.
The transitions can also be strongly asymmetric: for VLA-JEPA\cite{sun2026vla} at mild severity, $75\%$ of failures under the simultaneous condition occur on initial states that succeeded under every perturbation individually.

These patterns are not explained simply by policy stochasticity or nominal task difficulty.
Independent re-evaluation of stochastic policies preserves the emergent-failure and compensated-success rates within a few percentage points, while conditioning on nominally successful initial states leaves them nearly unchanged in almost all conditions.
Together, these results show that agreement at the level of aggregate success rates does not imply agreement at the level of matched outcomes.

Our contributions are:
\begin{itemize}
\item A fully paired six-axis LIBERO robustness protocol that evaluates individual and simultaneous perturbations on identical initial states with matched perturbation values across six comparable language-conditioned VLA policies.
\item A paired analysis that distinguishes overlap in single-axis vulnerabilities from two directional outcome transitions revealed when individual perturbations are compared with their simultaneous application: emergent failures and compensated successes.
\item Empirical evidence that opposing outcome transitions can substantially cancel: even when their aggregate difference is not statistically detectable, up to $29.0\%$ of matched outcomes change, while outcome changes reach $34.5\%$ overall. These patterns remain stable under controls for policy stochasticity and nominal task solvability.
\end{itemize}

\section{Related Work}
\label{sec:related}

\paragraph{Vision-language-action policies.}
Recent vision-language-action (VLA) research builds on scalable language-conditioned robot policies such as RT-1~\cite{brohan2022rt}, RT-2~\cite{brohan2023rt}, and OpenVLA~\cite{kim2024openvla}, and has since diversified substantially in model scale, pretraining strategy, action representation, and control formulation.
Large pretrained approaches include OpenVLA-OFT~\cite{kim2025fine}, which adapts OpenVLA with continuous action prediction, parallel decoding, and action chunking, and $\pi_{0.5}$~\cite{intelligence2025pi05}, which builds on the flow-matching $\pi_0$ family~\cite{black2024pi0}.
On the other hand, UniVLA~\cite{bu2025univla} learns task-centric latent actions from heterogeneous video, while SmolVLA~\cite{shukor2025smolvla} targets compact and efficient VLA deployment.
Predictive formulations have also emerged as a distinct direction in recent VLA research.
VLA-JEPA~\cite{sun2026vla} incorporates latent future-state prediction, whereas PredVLA~\cite{sawada2026predvla} adopts sub-million-parameter predictive-coding recurrent dynamics with online prediction-error inference.
These approaches span markedly different parameter scales, pretraining sources, representations, and action-generation mechanisms, making cross-family robustness analysis important rather than relying on conclusions from a single VLA formulation.

\paragraph{Manipulation benchmarks for language-conditioned policies.}
Robot manipulation policies have been studied across benchmarks with different emphases.
RLBench~\cite{james2020rlbench} provides a broad collection of visually grounded manipulation tasks, CALVIN~\cite{mees2022calvin} focuses on language-conditioned long-horizon task composition and generalization to novel environments, and LIBERO~\cite{liu2023libero} provides four suites for studying spatial, object, goal, and long-horizon manipulation.
More recent benchmarks such as RoboCasa~\cite{nasiriany2024robocasa}, VLABench~\cite{zhang2025vlabench}, and RoboCasa365~\cite{nasiriany2026robocasa365} further expand task, linguistic, and environmental diversity.

LIBERO has become a common evaluation setting for recent VLA policies, several of which report nominal aggregate success rates around or above $95\%$~\cite{intelligence2025pi05,kim2025fine,bu2025univla,sun2026vla}.
MINERVA~\cite{sendai2026minerva} further demonstrates $95.1\%$ nominal LIBERO success with a $0.54$M-parameter task-conditioned visuomotor policy.
Together, these results indicate that high nominal LIBERO success is attainable by policies with widely different scales and formulations, limiting the extent to which nominal task success alone characterizes policy robustness.

\paragraph{Robustness and generalization of VLA policies.}
Generalization benchmarks such as CALVIN~\cite{mees2022calvin}, VLABench~\cite{zhang2025vlabench}, and RoboCasa365~\cite{nasiriany2026robocasa365} evaluate policies across changes in tasks, environments, objects, or language.
More targeted robustness studies extend LIBERO with controlled perturbations.
LIBERO-Plus~\cite{liberoplus2025} evaluates seven perturbation dimensions and shows substantial degradation under changes such as camera viewpoint, robot initial state, language, lighting, background, sensor noise, and object layout.
It also studies selected pairwise perturbation compositions on OpenVLA-OFT~\cite{kim2025fine}, comparing observed joint performance with predictions derived from single-dimension success rates.
LIBERO-PRO~\cite{liberopro2025} similarly evaluates generalization under perturbations to objects, positions, instructions, tasks, and environments, including configurable combinations of perturbations.

These studies establish that nominal success does not imply robustness and that perturbation effects need not compose trivially at the aggregate level.
However, their reported analyses do not link the outcomes of all constituent perturbations to the corresponding compound outcome for the same evaluation instance under matched perturbation values.
Aggregate single-axis and compound success rates therefore cannot determine whether different perturbations fail on the same initial states, or whether an initial state that survives its constituent perturbations changes outcome when those perturbations are applied together.
The matched-instance relationship between constituent and compound robustness consequently remains undercharacterized.

\section{Method}
\label{sec:method}

\subsection{Evaluation protocol}
\label{sec:method:protocol}

Our main comparison includes six publicly released policies for which LIBERO-fine-tuned checkpoints are available and whose interfaces accept language instructions directly, allowing all six perturbation axes to be applied under a common evaluation protocol:
$\pi_{0.5}$~\cite{intelligence2025pi05},
OpenVLA-OFT~\cite{kim2025fine},
UniVLA~\cite{bu2025univla},
SmolVLA~\cite{shukor2025smolvla},
VLA-JEPA~\cite{sun2026vla},
and PredVLA~\cite{sawada2026predvla}.
These policies span sub-million to multi-billion parameter scales and substantially different pretraining and action-generation formulations.
MINERVA~\cite{sendai2026minerva} is evaluated separately as an auxiliary control.
Its interface maps instructions to fixed LIBERO task indices rather than processing free-form language, so the language intervention cannot be applied; its simultaneous condition therefore contains five effective perturbations and it is excluded from the main six-axis comparison.

We use all four LIBERO suites~\cite{liu2023libero}:
\textsc{Spatial}, \textsc{Object}, \textsc{Goal}, and \textsc{Long}.
Each suite contains 10 tasks with 50 released initial states, yielding 2,000 nominal rollouts per checkpoint.
Success is defined by simulator goal satisfaction within the episode budget; the budgets are 220, 280, 300, and 520 control steps for \textsc{Spatial}, \textsc{Object}, \textsc{Goal}, and \textsc{Long}, respectively.

LIBERO-CTRL evaluates each checkpoint under six single-axis perturbation conditions and one simultaneous condition at three severity levels.
The six axes intervene at different points in the observation--action loop:
\emph{camera} changes third-person camera pose and framing;
\emph{lighting} changes scene photometry;
\emph{robot} offsets the initial end-effector pose;
\emph{sensor} corrupts visual observations;
\emph{actuation} perturbs commanded-to-realised actions; and
\emph{language} applies meaning-preserving paraphrases.
For a given initial state and severity level, the simultaneous condition combines the same per-axis perturbation values used in the corresponding single-axis evaluations.

\begin{figure*}[!b]
\centering
\includegraphics[width=\textwidth]{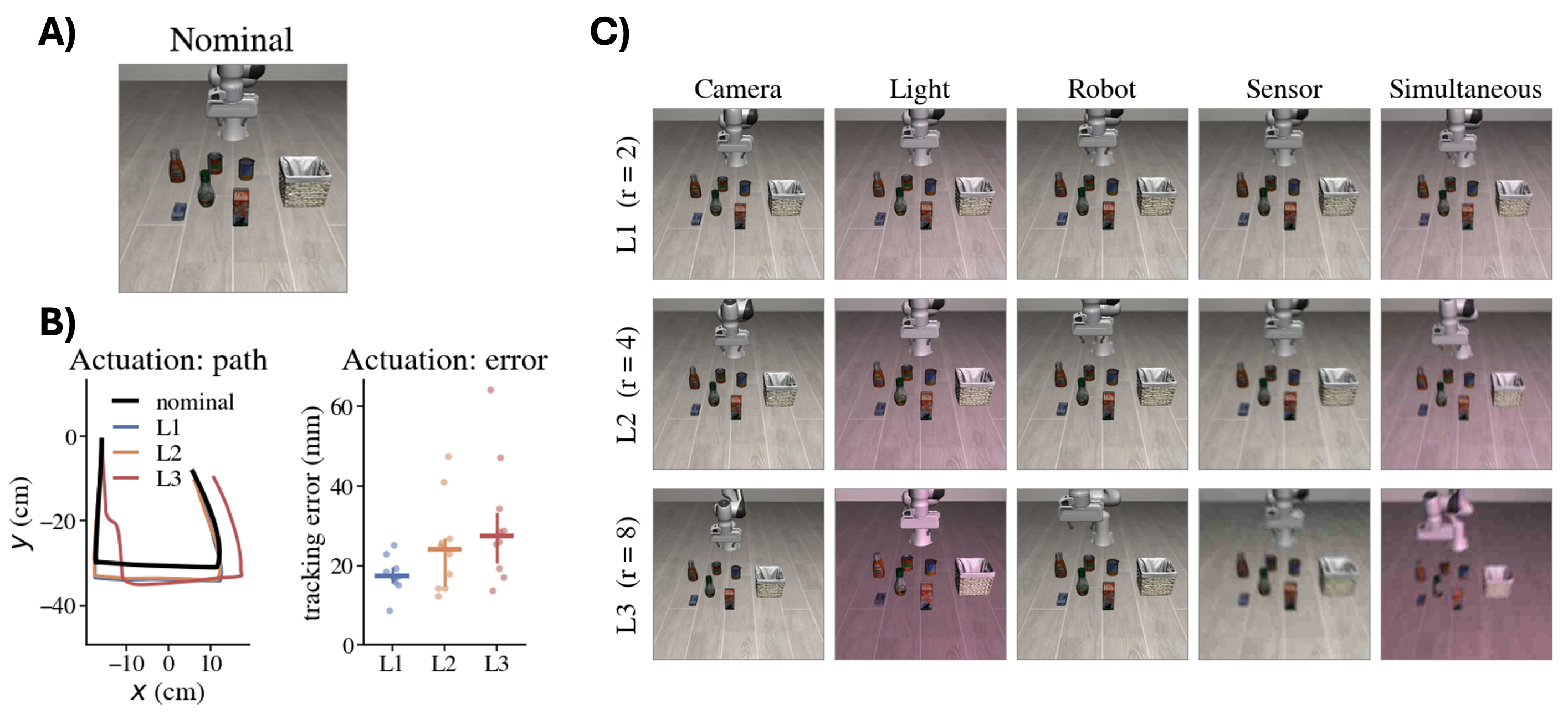}
\caption{
Illustration of LIBERO-CTRL perturbations for \textsc{Object} task~2,
configuration~0 (\emph{``pick up the salad dressing and place it in the basket''}).
\textbf{(A)} Nominal observation.
\textbf{(B)} Actuation perturbations, which do not alter the static visual observation.
The left plot shows the realised end-effector paths obtained by replaying the same open-loop command sequence under the nominal condition and under configuration~0 at $\mathrm{L}1$--$\mathrm{L}3$.
The right plot shows the time-averaged tracking error for all ten actuation presets at each severity level; points denote individual presets, with the median and interquartile range overlaid.
This panel is illustrative; the actuation normalisation scales themselves are calibrated separately on \textsc{Spatial} as described in Table~\ref{tab:params}.
\textbf{(C)} Camera, lighting, robot-pose, sensor, and simultaneous conditions for the same task and configuration.
For each metric axis, the ten presets at a given level lie at the same normalised radius $r$, while differing in direction in that axis's parameter space.
The simultaneous condition combines the matched per-axis presets for the same configuration.
Language is not visible in the RGB observation; for the illustrated configuration, the $\mathrm{L}3$ paraphrase is
\emph{``find the salad dressing, pick it up, and then place it into the basket''}.
}
\label{fig:perturbations}
\end{figure*}

For the perturbation analysis, we first construct a fixed evaluation set that is shared across all policies.
For each of the 40 LIBERO tasks, 10 of the 50 released initial states are randomly selected once and indexed by
$j\in\{0,\ldots,9\}$.
At each severity level, each perturbation axis likewise has 10 predefined perturbation presets, and preset $j$ is paired with initial state $j$.
This produces 10 fixed evaluation configurations per task, yielding 400 task--configuration pairs in total across the 40 LIBERO tasks.
The entire assignment of initial states and perturbation presets is generated before policy evaluation and held fixed across all models.

Each of these 400 configurations is evaluated under the six single-axis conditions and the simultaneous condition.
For a given task, configuration index, and severity level, the simultaneous rollout combines exactly the six per-axis perturbation presets used in the corresponding single-axis rollouts.
Thus, comparisons across conditions are matched both in initial state and in perturbation assignment.
Each severity level therefore contains $7 \times 400 = 2{,}800$ perturbed rollouts, and the three severity levels yield $8{,}400$ perturbed rollouts per checkpoint.
Together with the $2{,}000$ nominal rollouts over all released initial states, each primary checkpoint is evaluated over $10{,}400$ rollouts.
The six primary VLA checkpoints therefore contribute $62{,}400$ rollouts to the main comparison.
MINERVA follows the same rollout structure, with language treated as an identity intervention, and contributes an additional $10{,}400$ auxiliary rollouts.
Independent re-evaluations used to assess policy stochasticity are reported separately and are not included in these totals.

\subsection{Perturbation construction and severity}
\label{sec:method:paired}

For each metric perturbation axis, parameter offsets are represented in normalised coordinates
$u_i=\Delta p_i/\sigma_i$, where $\sigma_i$ is a fixed parameter-specific scale.
Severity is defined by $r=\lVert\mathbf{u}\rVert_2$, with
$\mathrm{L}1$, $\mathrm{L}2$, and $\mathrm{L}3$ corresponding to
$r=2$, $4$, and $8$, respectively.
Within each metric axis and severity level, the ten perturbation presets differ in direction while sharing the same normalised magnitude.
Camera, lighting, and sensor scales are anchored to predefined physical ranges,
robot-pose scales to the spacing between released initial states,
and actuation scales to measured open-loop end-effector tracking error.
Language is categorical rather than metrically calibrated.
Figure~\ref{fig:perturbations} illustrates the resulting perturbations;
Table~\ref{tab:params} reports the complete parameter-wise normalisation scales.

\begin{table}[t]
\centering
\caption{Perturbation normalisation scales.
The final column shows the displacement at $r=8$ when allocated entirely to one parameter.}
\label{tab:params}
\footnotesize
\begin{tabular}{@{}llrr@{}}
\toprule
Axis & Parameter & $\sigma$ & at $r=8$ \\
\midrule
\multirow{5}{*}{Camera}
 & azimuth & $3.48^\circ$ & $27.9^\circ$ \\
 & elevation & $4.77^\circ$ & $38.2^\circ$ \\
 & distance & $6.4$\,cm & $51.2$\,cm \\
 & look-at $x$ & $1.95$\,cm & $15.6$\,cm \\
 & look-at $y$ & $1.33$\,cm & $10.6$\,cm \\
\midrule
\multirow{5}{*}{Lighting}
 & intensity & $0.25$\,EV & $2.0$\,EV ($\times4$) \\
 & warmth (R:B) & $0.19\,\log_2$ & $1.5\,\log_2$ ($\times2.8$) \\
 & green tint & $0.10\,\log_2$ & $0.8\,\log_2$ ($\times1.7$) \\
 & ambient & $0.25\,\log_2$ & $2.0\,\log_2$ ($\times4$) \\
 & light elevation & $7.5^\circ$ & $60^\circ$ \\
\midrule
\multirow{2}{*}{Robot pose$^{\dagger}$}
 & EEF position & $2.57$\,cm & $20.6$\,cm \\
 & EEF rotation & $3.94^\circ$ & $31.5^\circ$ \\
\midrule
\multirow{4}{*}{Sensor}
 & pixel noise $\sigma$ & $2.6/255$ & $20.4/255$ \\
 & Gaussian blur & $0.25$\,px & $2.0$\,px \\
 & JPEG quality & $0.42\,\log_2$ & $3.3\,\log_2$ ($100\!\to\!10$) \\
 & motion blur length & $0.75$\,px & $6.0$\,px \\
\midrule
\multirow{5}{*}{Actuation$^{\ddagger}$}
 & gain & $0.014\,\log_2$ & $0.115\,\log_2$ ($+8.3\%$) \\
 & bias & $0.74\%$ & $5.95\%$ \\
 & misalignment & $0.90^\circ$ & $7.17^\circ$ \\
 & lag & $16$\,ms & $130$\,ms \\
 & actuator noise & $2.35\%$ & $18.8\%$ \\
\midrule
Language$^{\S}$ &
\multicolumn{3}{l}{categorical; nested $\mathrm{L}1\!\subset\!\mathrm{L}2\!\subset\!\mathrm{L}3$} \\
\bottomrule
\end{tabular}

\footnotesize
$^{\dagger}$$\sigma$ is derived from the spacing among a task's $50$ released initial end-effector states.
$^{\ddagger}$Actuation scales are calibrated on \textsc{Spatial} so that
$r=2,4,8$ produce approximately $10/20/40$\,mm of open-loop end-effector tracking error.
$^{\S}$Language $\mathrm{L}1$ changes surface vocabulary,
$\mathrm{L}2$ additionally rewrites local structure,
and $\mathrm{L}3$ additionally rewrites sentence frames.
\end{table}

\subsection{Single-axis marginals and paired conjunction}
\label{sec:method:models}

At a fixed severity level, let $x\in\{1,\ldots,N\}$ index the $N=400$ matched evaluation configurations.
Let $Y_a(x)\in\{0,1\}$ denote the binary outcome for configuration $x$ when perturbation axis $a$ is applied alone, and let $Y_{\mathrm{sim}}(x)$ denote the outcome when all six matched perturbations are applied simultaneously.
We define the marginal single-axis and simultaneous success rates as $S_a=\mathbb{E}[Y_a]$ and $S_{\mathrm{sim}}=\mathbb{E}[Y_{\mathrm{sim}}]$, respectively.
From the single-axis outcomes, we further define
\begin{equation}
S_{\mathrm{indep}}
=\prod_{a=1}^{6}S_a,\qquad
S_{\mathrm{conj}}
=\mathbb{E}\!\left[\prod_{a=1}^{6}Y_a(x)\right].
\label{eq:single_axis}
\end{equation}

\begin{figure*}[t]
\centering
\includegraphics[width=\textwidth]{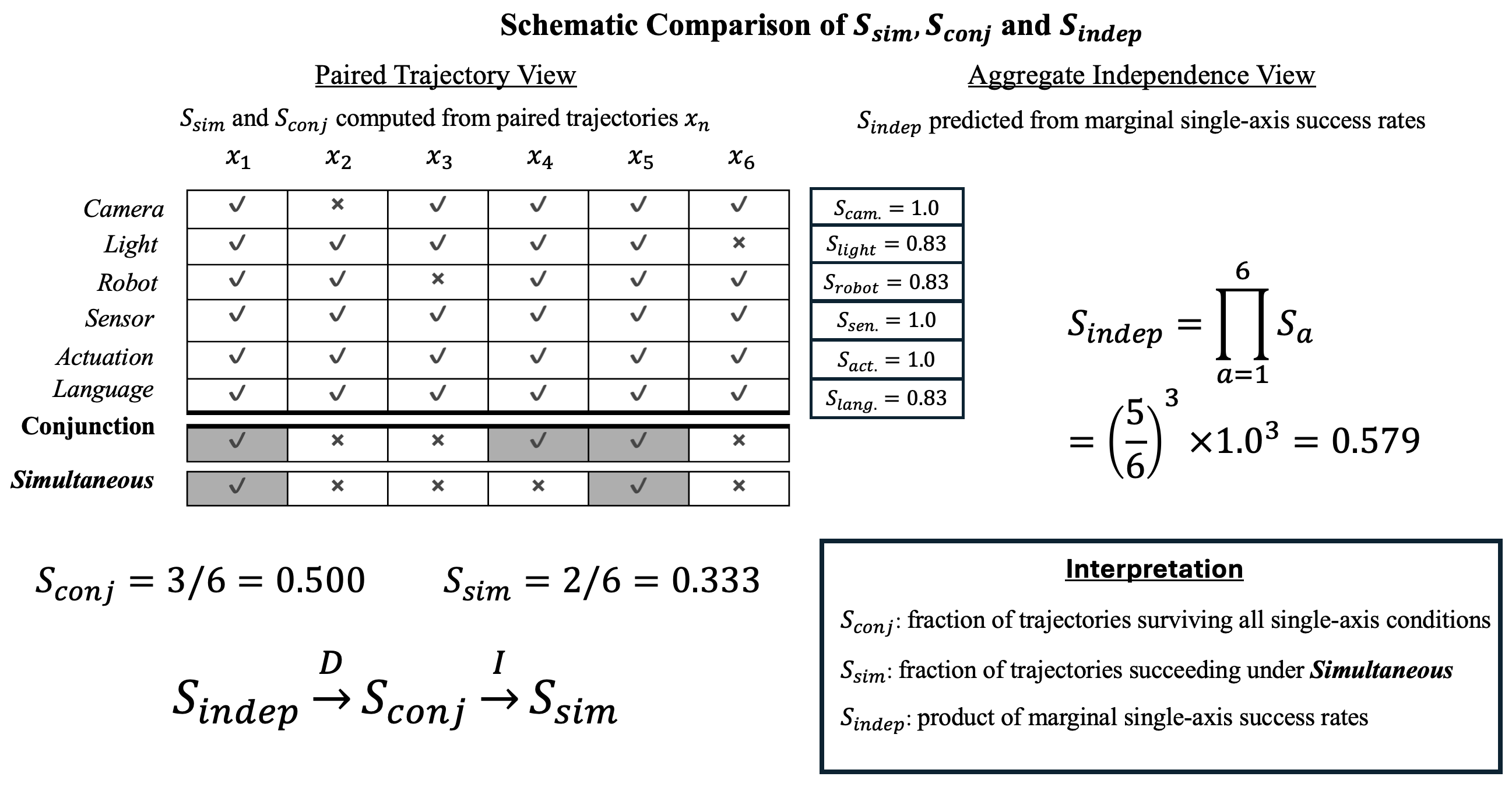}
\caption{
Schematic illustration of the success measures.
$S_{\mathrm{indep}}$ is obtained from the marginal single-axis success rates, $S_{\mathrm{conj}}$ is the observed fraction of matched configurations that succeed under all six single-axis conditions individually, and $S_{\mathrm{sim}}$ is the observed success rate when the matched perturbations are applied simultaneously.
The example is conceptual and does not represent experimental data.
}
\label{fig:schematic_s}
\end{figure*}

$S_{\mathrm{indep}}$ is the conjunction success rate expected if the six single-axis success events were statistically independent.
In contrast, $S_{\mathrm{conj}}$ is observed directly from the paired evaluations and is the fraction of matched configurations that succeed under every perturbation individually.
The difference $S_{\mathrm{conj}}-S_{\mathrm{indep}}$ therefore captures dependence in which configurations are vulnerable across single-axis perturbations.
Figure~\ref{fig:schematic_s} summarises the relationship among $S_{\mathrm{indep}}$, $S_{\mathrm{conj}}$, and $S_{\mathrm{sim}}$.

\subsection{Decomposition and matched outcome transitions}
\label{sec:method:decomp}

The difference between simultaneous success and the independence prediction decomposes exactly as
\begin{equation}
\begin{aligned}
S_{\mathrm{sim}}-S_{\mathrm{indep}}
&=
\underbrace{(S_{\mathrm{conj}}-S_{\mathrm{indep}})}_{D}
+
\underbrace{(S_{\mathrm{sim}}-S_{\mathrm{conj}})}_{I}.
\end{aligned}
\label{eq:decomp}
\end{equation}

The term $D$ captures cross-axis dependence already present in the paired single-axis outcomes.
The remaining term, $I=S_{\mathrm{sim}}-S_{\mathrm{conj}}$, is the signed difference in success rate between the simultaneous condition and the paired single-axis conjunction.
Thus, $D$ can be estimated entirely from the single-axis evaluations, whereas $I$ requires the simultaneous condition.

A signed difference alone can, however, conceal changes in opposite directions.
Let $b$ denote the number of matched configurations that succeed under all six single-axis conditions but fail under the simultaneous condition, and let $c$ denote the number that fail under at least one single-axis condition but succeed under the simultaneous condition.
For $N=400$ configurations, we define
\begin{equation}
\begin{gathered}
R_{\mathrm e}=\frac{b}{N},\qquad
R_{\mathrm c}=\frac{c}{N},\\
R_{\mathrm d}=R_{\mathrm e}+R_{\mathrm c},\qquad
I=R_{\mathrm c}-R_{\mathrm e}.
\end{gathered}
\label{eq:paired_metrics}
\end{equation}

Here, $R_{\mathrm e}$ is the \emph{emergent-failure rate}, where a configuration succeeds under every constituent perturbation individually but fails when they are combined; $R_{\mathrm c}$ is the \emph{compensated-success rate}, where at least one constituent condition fails but the simultaneous condition succeeds; and $R_{\mathrm d}$ is their total disagreement rate.
The identity $I=R_{\mathrm c}-R_{\mathrm e}=S_{\mathrm{sim}}-S_{\mathrm{conj}}$ shows that emergent failures and compensated successes contribute with opposite signs to the aggregate difference and can therefore cancel.

\subsection{Statistical inference and stochasticity control}
\label{sec:method:controls}

At each severity level, the $400$ matched configurations comprise $40$ task clusters with $10$ configurations per task.
We compute $95\%$ confidence intervals using a cluster bootstrap that resamples the $40$ tasks with replacement while retaining all configurations within each sampled task, using $4{,}000$ bootstrap replicates.

To assess sensitivity to policy stochasticity, we independently repeat the complete $8{,}400$-rollout perturbation evaluation for the three policies with stochastic inference, $\pi_{0.5}$, SmolVLA, and VLA-JEPA, using a different random seed while keeping the initial states and perturbation assignments fixed.
We then recompute $R_{\mathrm e}$, $R_{\mathrm c}$, $R_{\mathrm d}$, and $I$ from each independent evaluation.
These re-evaluations serve as replication controls and are excluded from the main benchmark totals.

\subsection{Nominal reproduction check}
\label{sec:method:repro}

Suite-specific checkpoints are used where provided.
Before interpreting robustness results, we verify each released checkpoint against its published aggregate nominal performance.
All checkpoints except SmolVLA reproduce the corresponding published score within $1.9$ percentage points.
SmolVLA~\cite{shukor2025smolvla} is treated separately because its publicly released LIBERO checkpoint was trained with one eighth of the sample budget used for the published result.
Its measured nominal score is therefore used as the checkpoint-specific reference for subsequent robustness analysis.

All evaluation code, policy adapters, perturbation configurations, rollout manifests, and per-rollout outcomes are available in the anonymous repository.\footnote{\url{https://github.com/hiroki-oist/LIBERO-ctrl}}

\section{Results}
\label{sec:results}

\subsection{Reproduction and single-axis robustness}
\label{sec:res:repro}

Six policies reproduce the corresponding published aggregate score within $1.9$ percentage points, as shown in Table~\ref{tab:clean}.
SmolVLA is a declared exception because its released checkpoint was trained with one eighth of the sample budget used for the published result.
MINERVA reproduces its published score but is treated only as an auxiliary identity-language control.

\begin{table}[t]
\centering
\caption{Nominal LIBERO success rate (\%) over $2{,}000$ rollouts per policy.}
\label{tab:clean}
\small
\begin{tabular}{lrrr}
\toprule
Policy & Params & Ours & Published \\
\midrule
$\pi_{0.5}$~\cite{intelligence2025pi05} & 4.14\,B & 96.20 & 96.9 \\
OpenVLA-OFT~\cite{kim2025fine} & 7.54\,B & 96.70 & 97.1 \\
UniVLA~\cite{bu2025univla} & 7.54\,B & 93.90 & 95.2 \\
SmolVLA~\cite{shukor2025smolvla} & 450\,M & 76.35 & 87.3$^\dagger$ \\
VLA-JEPA~\cite{sun2026vla} & 2.77\,B & 97.60 & 97.2 \\
PredVLA~\cite{sawada2026predvla} & 0.68\,M & 79.30 & 77.80 \\
MINERVA~\cite{sendai2026minerva}$^{\ast}$ & 0.54\,M & 93.90 & 95.75 \\
\bottomrule
\end{tabular}

\vspace{2pt}
\footnotesize
$^{\dagger}$Released checkpoint uses one eighth of the published training sample budget.
$^{\ast}$Auxiliary identity-language control.
\end{table}

\begin{figure*}[b]
\centering
\includegraphics[width=\textwidth]{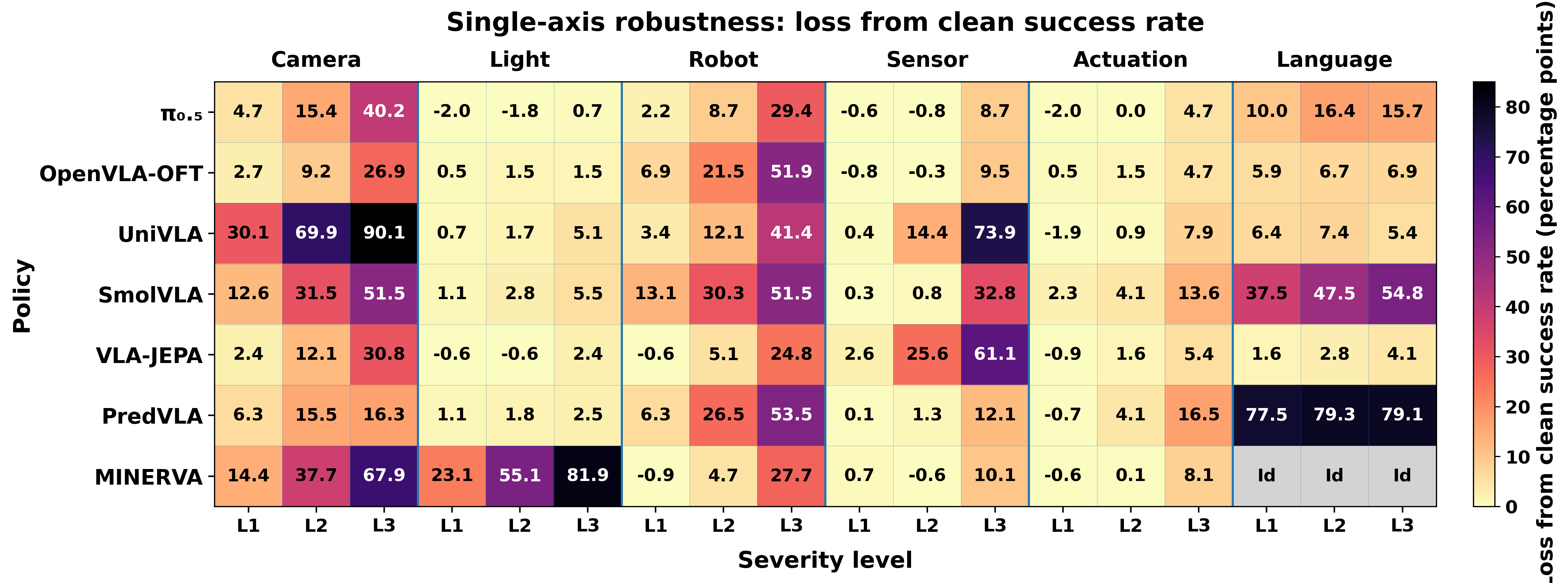}
\caption{
Single-axis robustness shown as clean-relative loss in success rate.
Rows are policies and columns are perturbation axes at severity levels $\mathrm{L}1$--$\mathrm{L}3$; darker cells indicate larger degradation.
MINERVA's language entries are identity interventions.
}
\label{fig:single_axis}
\end{figure*}

All policies show strongly anisotropic robustness profiles.
For example, UniVLA falls to $3.8\%$ under camera perturbation at $\mathrm{L}3$ while retaining $88.8\%$ under lighting, whereas PredVLA remains comparatively robust to several metric perturbations but collapses under language paraphrasing.
Figure~\ref{fig:single_axis} summarises the complete single-axis profiles.

\subsection{Cross-axis dependence and matched outcome transitions}
\label{sec:res:paired}

\begin{figure*}[t]
\centering
\includegraphics[width=\textwidth]{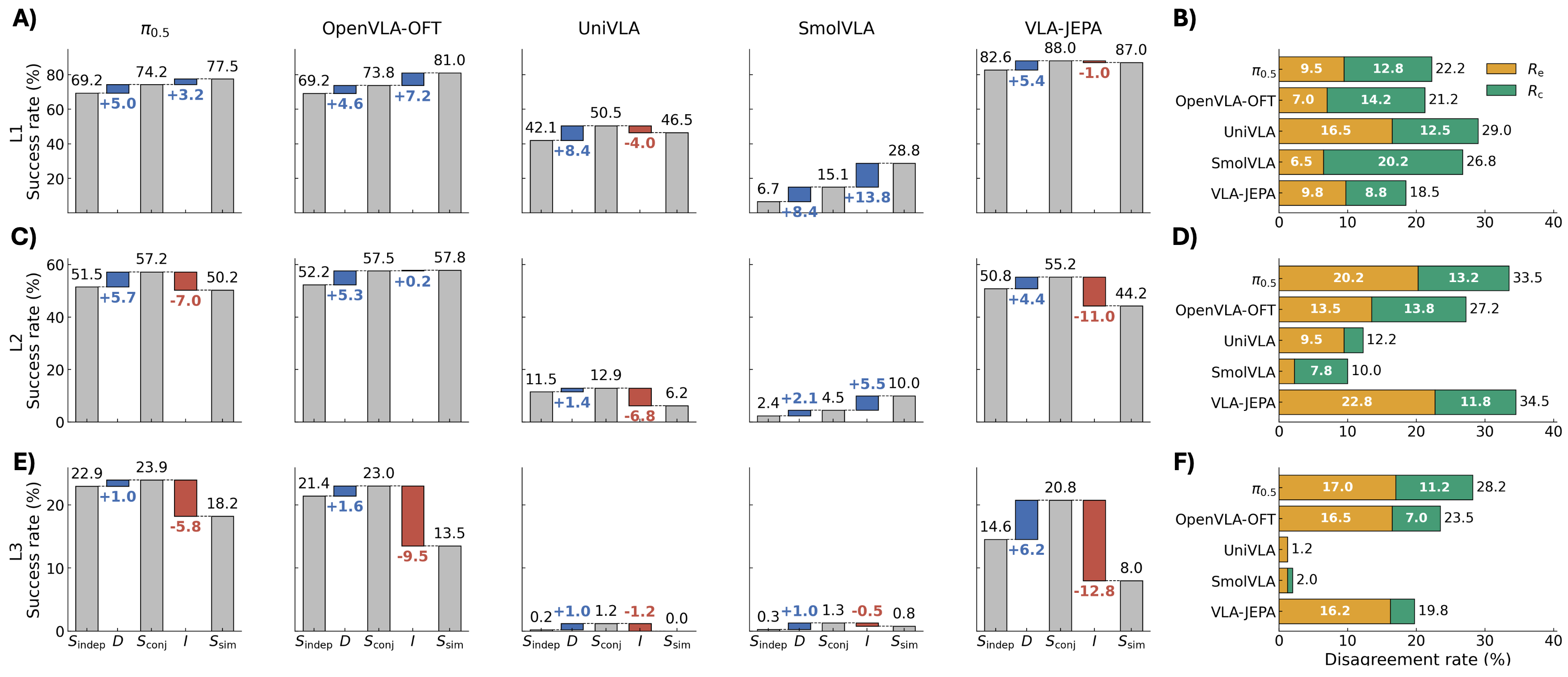}
\caption{
Compound robustness decomposition across five policies and three severity levels.
\textbf{(A,B)} $\mathrm{L}1$, \textbf{(C,D)} $\mathrm{L}2$, and \textbf{(E,F)} $\mathrm{L}3$.
\textbf{(A,C,E)} $S_{\mathrm{indep}}$, $S_{\mathrm{conj}}$, and $S_{\mathrm{sim}}$ are shown as absolute success rates in grey, while $D=S_{\mathrm{conj}}-S_{\mathrm{indep}}$ and $I=S_{\mathrm{sim}}-S_{\mathrm{conj}}$ show the corresponding changes between them.
Positive changes are shown in blue and negative changes in red.
Within each severity level, all policies share the same vertical scale.
\textbf{(B,D,F)} The total disagreement rate $R_{\mathrm d}$ is decomposed into emergent failures $R_{\mathrm e}$ and compensated successes $R_{\mathrm c}$, shown in orange and teal, respectively, revealing opposing outcome transitions that may cancel in the signed aggregate effect $I$.
}
\label{fig:overall_results}
\end{figure*}

The paired conjunction $S_{\mathrm{conj}}$ differs from the independence prediction $S_{\mathrm{indep}}$, indicating that vulnerability to individual perturbations is not distributed independently across matched configurations.
Across the five policies visualised in Fig.~\ref{fig:overall_results}, $D=S_{\mathrm{conj}}-S_{\mathrm{indep}}$ is positive at all three severity levels.
For example, OpenVLA-OFT at $\mathrm{L}1$ has $D=+4.6$ percentage points.
Thus, marginal single-axis success rates alone do not determine how often the same configurations survive all constituent perturbations individually.

More importantly, the signed difference $I=S_{\mathrm{sim}}-S_{\mathrm{conj}}$ can conceal substantial outcome changes in opposite directions.
Figure~\ref{fig:overall_results} visualises both the aggregate decomposition and the directional transition rates.
OpenVLA-OFT at $\mathrm{L}2$ provides a clear example: $R_{\mathrm e}=13.50\%$ and $R_{\mathrm c}=13.75\%$, yielding $R_{\mathrm d}=27.25\%$ but only $I=+0.25$ percentage points, with a $95\%$ task-cluster-bootstrap CI of $[-6.00,+6.51]$ points.
Thus, more than one quarter of matched configurations change binary outcome between the single-axis conjunction and the simultaneous condition even though the two opposing transitions almost exactly cancel in aggregate.

This cancellation is not unique to OpenVLA-OFT.
At $\mathrm{L}1$, UniVLA, $\pi_{0.5}$, and VLA-JEPA have disagreement rates of $29.00\%$, $22.25\%$, and $18.50\%$, respectively, while the $95\%$ confidence interval for $I$ includes zero in each case.
UniVLA $\mathrm{L}1$ gives the largest disagreement observed among conditions whose confidence interval for $I$ includes zero, with $I=-4.00$ points, $95\%$ CI $[-10.00,+2.25]$, and $R_{\mathrm d}=29.00\%$.
These cases show that an aggregate change that is not distinguishable from zero can coexist with substantial redistribution of which matched configurations succeed and fail.

Across all evaluated policy--severity conditions, the largest disagreement is $34.50\%$ for VLA-JEPA at $\mathrm{L}2$.
Unlike the cancellation cases above, this condition also exhibits a clear directional imbalance, with $I=-11.00$ points and a $95\%$ CI of $[-19.50,-2.75]$.
The second-largest disagreement is $33.50\%$ for $\pi_{0.5}$ at $\mathrm{L}2$, where $I=-7.00$ points with a $95\%$ CI of $[-13.50,-0.50]$.
Across the $18$ main policy--severity cells, the $95\%$ confidence interval for $I$ excludes zero in $10$.
Together, these results distinguish two cases that aggregate success rates alone conflate: substantial bidirectional changes that largely cancel, and substantial changes with a persistent imbalance toward one transition direction.

The directional decomposition also clarifies how much of simultaneous failure is genuinely emergent.
At VLA-JEPA $\mathrm{L}1$, $R_{\mathrm e}=9.75\%$, while the simultaneous failure rate is $13.0\%$.
Emergent failures therefore account for $75.0\%$ of all simultaneous-condition failures in this cell.
In other words, most compound failures occur on matched configurations that succeeded under every constituent perturbation individually.

PredVLA lies near the compound-performance boundary and is omitted from Fig.~\ref{fig:overall_results} for visual clarity.
Its simultaneous success rates are $1.2\%$, $0.0\%$, and $0.2\%$ at $\mathrm{L}1$--$\mathrm{L}3$, respectively, while $R_{\mathrm d}$ reaches at most $1.25\%$.
Accordingly, the paired analysis has little remaining success probability with which to resolve additional compound degradation for this policy.

\subsection{Stochasticity replication}
\label{sec:res:controls}

Independent stochastic-policy replications change $R_{\mathrm e}$ by at most $2.25$ percentage points and $R_{\mathrm c}$ by at most $2.75$ points.
For $\pi_{0.5}$ at $\mathrm{L}2$, the original evaluation gives $I=-7.00$ points with a $95\%$ CI of $[-13.50,-0.50]$, whereas the independent replication shifts $I$ to $-3.00$ points.
Its directional components remain similar, with $R_{\mathrm e}$ changing from $20.25\%$ to $18.50\%$ and $R_{\mathrm c}$ from $13.25\%$ to $15.50\%$.
SmolVLA $\mathrm{L}2$ similarly no longer yields a confidence interval for $I$ excluding zero in the independent replication, despite an original estimate of $I=+5.50$ points with $95\%$ CI $[+1.75,+9.50]$.
In contrast, the larger directional imbalances for SmolVLA $\mathrm{L}1$, VLA-JEPA $\mathrm{L}2$--$\mathrm{L}3$, and $\pi_{0.5}$ $\mathrm{L}3$ remain supported under replication.
Overall, the replication indicates that the large emergent and compensated components are stable, while inference about their smaller signed difference $I$ is more sensitive to stochastic policy variation.

\section{Discussion}
\label{sec:discussion}

\subsection{Interpretation of compound robustness}

Single-axis robustness and robustness under compound perturbations are distinct properties.
The difference between $S_{\mathrm{indep}}$ and $S_{\mathrm{conj}}$ shows that marginal success rates do not capture which configurations are vulnerable across perturbation axes, while $I=S_{\mathrm{sim}}-S_{\mathrm{conj}}$ captures the additional change observed when those perturbations are applied together.
These quantities therefore separate dependence already present in the single-axis outcomes from changes specific to their simultaneous application.

However, $I$ is itself an aggregate quantity.
Because $I=R_{\mathrm c}-R_{\mathrm e}$, emergent failures and compensated successes can cancel, allowing a near-zero signed difference despite substantial matched-outcome disagreement.
A small $I$ should therefore be interpreted as a balance between directional transitions, not necessarily as agreement between the single-axis conjunction and simultaneous outcomes.

The practical implication is that compound robustness should be evaluated using matched outcomes rather than aggregate success rates alone.
A paired single-axis sweep provides $S_{\mathrm{conj}}$ and $D$, while one additional matched \emph{Simultaneous} condition identifies $R_{\mathrm e}$, $R_{\mathrm c}$, $R_{\mathrm d}$, and $I$.
Reporting the directional rates avoids losing substantial outcome changes through aggregation.

\subsection{Limits of single-axis evaluation}

Single-axis robustness is strongly anisotropic across policies and perturbation axes.
A low compound success rate may therefore be inherited from a dominant single-axis vulnerability rather than from an additional effect of combining perturbations.
PredVLA illustrates this boundary case, where language perturbation already drives performance close to zero and leaves limited resolution for further negative compound effects.

Conversely, strong constituent performance does not guarantee compound reliability.
Emergent failures occur on configurations that succeed under every perturbation individually, and therefore cannot be identified from marginal single-axis success rates.
Compensated successes further show that compound outcomes are not simply a stricter version of the constituent conditions.

The prevalence and direction of these transitions vary substantially across policies.
Because the evaluated checkpoints differ simultaneously in architecture, pretraining, data, and action-generation formulation, these cross-policy differences should not be interpreted mechanistically.
The stochastic-policy replications nevertheless preserve the directional components more consistently than their signed difference, further supporting separate reporting of $R_{\mathrm e}$ and $R_{\mathrm c}$.

\subsection{Scope and implications}

Our results concern controlled compound perturbations in simulation rather than arbitrary real-world distribution shift.
The \emph{Simultaneous} condition combines all six tested axes and therefore cannot identify which lower-order subsets produce a particular transition; targeted combinatorial evaluation would be required to localise such effects.
The language axis is categorical rather than metrically calibrated, and MINERVA remains an auxiliary control because its interface does not admit the language perturbation.

The directional rates are also constrained near performance boundaries.
When $S_{\mathrm{conj}}$ is close to zero, few additional emergent failures are possible, while a very low $S_{\mathrm{sim}}$ limits compensated successes.
Near-boundary cases should therefore not be interpreted as strong evidence that compound effects are absent.

Within these limits, neither marginal single-axis success rates nor the signed difference $S_{\mathrm{sim}}-S_{\mathrm{conj}}$ fully characterises compound robustness.
Pairing evaluations separates shared constituent vulnerability from outcome changes that emerge under simultaneous perturbation, and the directional decomposition reveals whether those changes reflect emergent failures, compensated successes, or both.

\section{Conclusion}
\label{sec:conclusion}

We introduced LIBERO-CTRL, a fully paired six-axis perturbation protocol for evaluating whether compound robustness can be inferred from single-axis measurements.
Our main analysis covers six language-conditioned policies, with MINERVA retained as an auxiliary identity-language control.
The paired design separates cross-axis dependence already present in single-axis outcomes from the additional signed change observed when perturbations are applied simultaneously.

The central finding is that the signed difference $I=S_{\mathrm{sim}}-S_{\mathrm{conj}}$ can conceal substantial matched-outcome changes because emergent failures and compensated successes contribute with opposite signs.
Thus, neither marginal single-axis success rates nor $I$ alone fully characterises compound robustness.
Paired evaluation should therefore report the two directional transition rates separately.
Given an already paired single-axis sweep, this requires only one additional matched \emph{Simultaneous} condition.

\paragraph{Limitations}
Our study evaluates released checkpoints in simulation on a single embodiment and cannot attribute cross-policy differences to specific architectures or training procedures.
The \emph{Simultaneous} condition does not identify which lower-order perturbation combinations produce each transition, and near-boundary conditions provide limited resolution in one transition direction.
The language axis is categorical rather than metrically calibrated, so conclusions about graded perturbation magnitude apply only to the metric axes.

\bibliographystyle{IEEEtran}
\bibliography{references}

\end{document}